\documentclass[10pt,twocolumn,letterpaper]{article}

\usepackage{iccv}
\usepackage{times}
\usepackage{epsfig}
\usepackage{graphicx}
\usepackage{amsmath}
\usepackage{cite}
\usepackage{amssymb}
\usepackage{multirow}
\usepackage{diagbox}
\usepackage{array}
\usepackage{subfigure}
\usepackage{overpic}
\usepackage{enumitem}

\pdfoutput=1
\usepackage[pagebackref=false,breaklinks=true,letterpaper=true,colorlinks,bookmarks=false]{hyperref}

\iccvfinalcopy % *** Uncomment this line for the final submission

\def\iccvPaperID{1164} % *** Enter the ICCV Paper ID here

\ificcvfinal\fi

\newcommand{\ST}{ST\cite{2014tip/ST}}
\newcommand{\DRFI}{DRFI\cite{WangDRFI2017}}
\newcommand{\DSR}{DSR\cite{2013iccv/DSR}}

\newcommand{\MC}{MC\cite{2015cvpr/MC}}
\newcommand{\MDF}{MDF\cite{2015cvpr/MDF}}
\newcommand{\DISC}{DISC\cite{2016nnls/DISC}}
\newcommand{\DCL}{DCL\cite{li2016deep/DCL}}
\newcommand{\rfcn}{rfcn\cite{2016eccv/rfcn}}
\newcommand{\DHS}{DHS\cite{2016cvpr/DHS}}
\newcommand{\ELD}{ELD\cite{lee2016deep/ELD}}

\newcommand{\CA}{CA\cite{2012pami/CA}}
\newcommand{\CB}{CB\cite{2012bmvc/CB}}
\newcommand{\RC}{RC\cite{cheng2015global}}
\newcommand{\PCA}{PCA\cite{2013cvpr/PCA}}
\newcommand{\SVO}{SVO\cite{2011iccv/SVO}}

\newcommand{\PASCAL}{PASCAL-S\cite{2014cvpr/PASCAL-S}}
\newcommand{\ECSSD}{ECSSD\cite{2013tip/ECSSD}}
\newcommand{\HKU}{HKU-IS\cite{2015cvpr/MDF}}
\newcommand{\SOD}{SOD\cite{2001iccv/SOD}}

\newcommand{\figref}[1]{Fig.~\ref{#1}}
\newcommand{\tabref}[1]{Tab.~\ref{#1}}
\newcommand{\equref}[1]{Equ.~(\ref{#1})}
\newcommand{\secref}[1]{Sec. \ref{#1}}

\newcommand{\sArt}{state-of-the-art }

\graphicspath{{./}}

\begin{document}

%%%%%%%%% TITLE
\title{Structure-measure: A New Way to Evaluate Foreground Maps}

\author{Deng-Ping Fan$^1$ \quad \quad  Ming-Ming Cheng$^1$ \thanks{M.M. Cheng (cmm@nankai.edu.cn) is the corresponding author.}
    \quad \quad Yun Liu$^1$ \quad \quad Tao Li$^1$
    \quad \quad Ali Borji$^2$ \\
    $^1$ CCCE, Nankai University \quad \quad $^2$ CRCV, UCF \\
    {\tt\small http://dpfan.net/smeasure/}
}

\maketitle
\thispagestyle{empty}

%%%%%%%%% ABSTRACT
\begin{abstract}
Foreground map evaluation is crucial for gauging the progress of object
segmentation algorithms, in particular in the field of salient object
detection where the purpose is to accurately detect and segment the most
salient object in a scene. Several widely-used measures such as Area Under
the Curve (AUC), Average Precision (AP) and the recently proposed
$F_{\beta}^{\omega}$(Fbw) have been used to evaluate the similarity
between a non-binary saliency map (SM) and a ground-truth (GT) map.
These measures are based on pixel-wise errors and often ignore the structural
similarities. Behavioral vision studies, however, have shown
that the human visual system is highly sensitive to structures in scenes.
Here, we propose a novel, efficient, and easy to calculate measure known
as structural similarity measure ($\textbf{Structure-measure}$) to evaluate
non-binary foreground maps.
Our new measure simultaneously evaluates region-aware and
object-aware structural similarity between a SM and a GT map.
We demonstrate superiority of our measure over existing ones using 5
meta-measures on 5 benchmark datasets.

\end{abstract}

\begin{figure}[t!]
  \centering
  \begin{overpic}[width=.95\columnwidth]{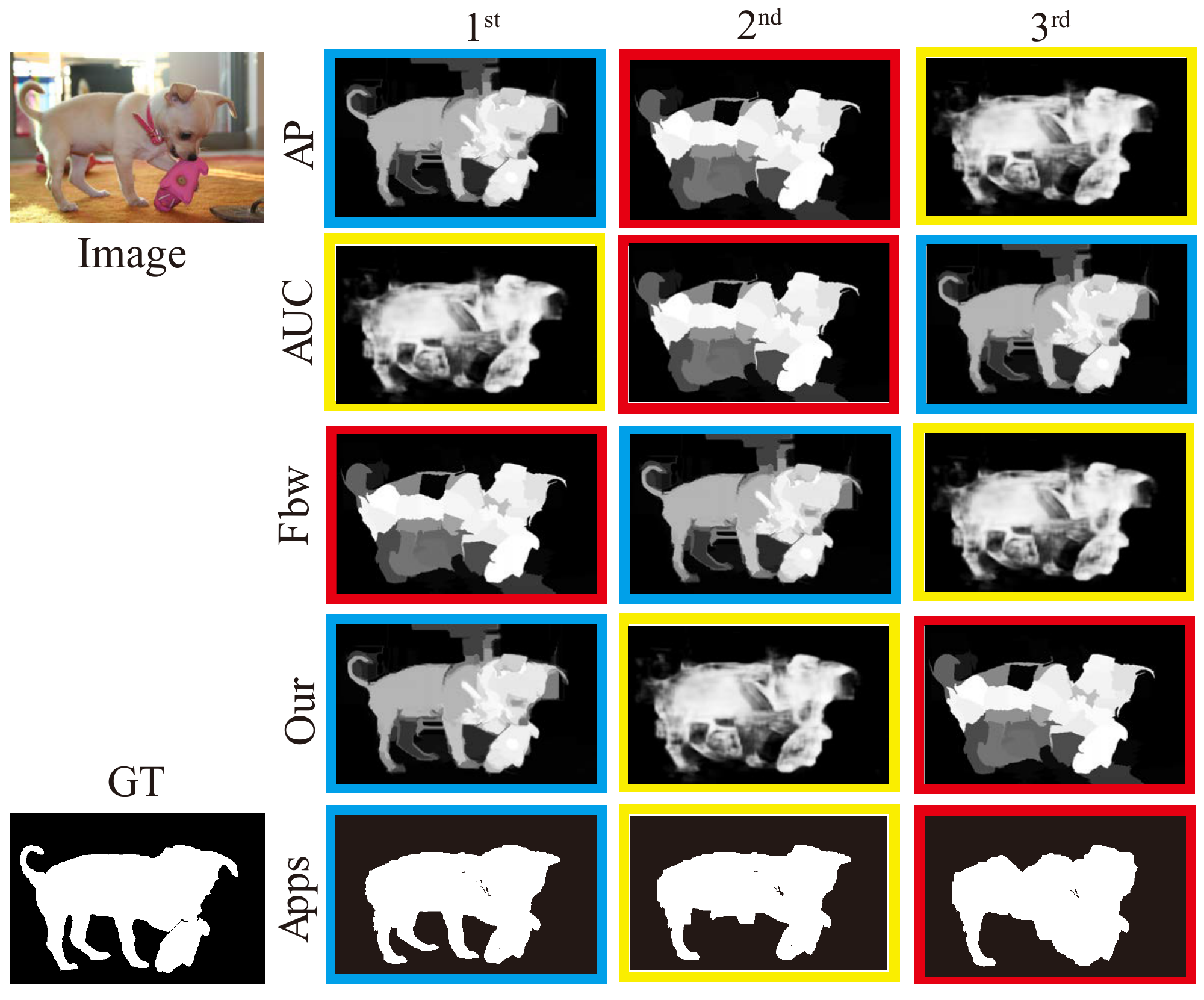}
  \end{overpic}
  \caption{\small\textbf{Inaccuracy of existing evaluation measures.}
    We compare the ranking of saliency maps generated by 3
    \sArt salient object detection algorithms: \DISC, \MDF, and \MC.
    According to the application's
    ranking (last row; \secref{sec:ExperimentsApps}),
    the blue-border map ranks first,
    followed by the yellow- and red-border maps.
    The blue-border map captures the dog's structure most accurately,
    with respect to the GT.
    The yellow-border map looks fuzzy although the overall outline of
    the dog is still present.
    The red-border map almost completely destroyed the structure
    of the dog.
    Surprisingly, all of the measures based on pixel-wise errors
    (first 3 rows) fail to rank the maps correctly.
    Our new measure (4th row) ranks the three maps in the right order.
  }\label{fig:FirsCmpDog}
  \vspace{-10pt}
\end{figure}

\begin{figure*}[t!]
  \begin{overpic}[width=\textwidth]{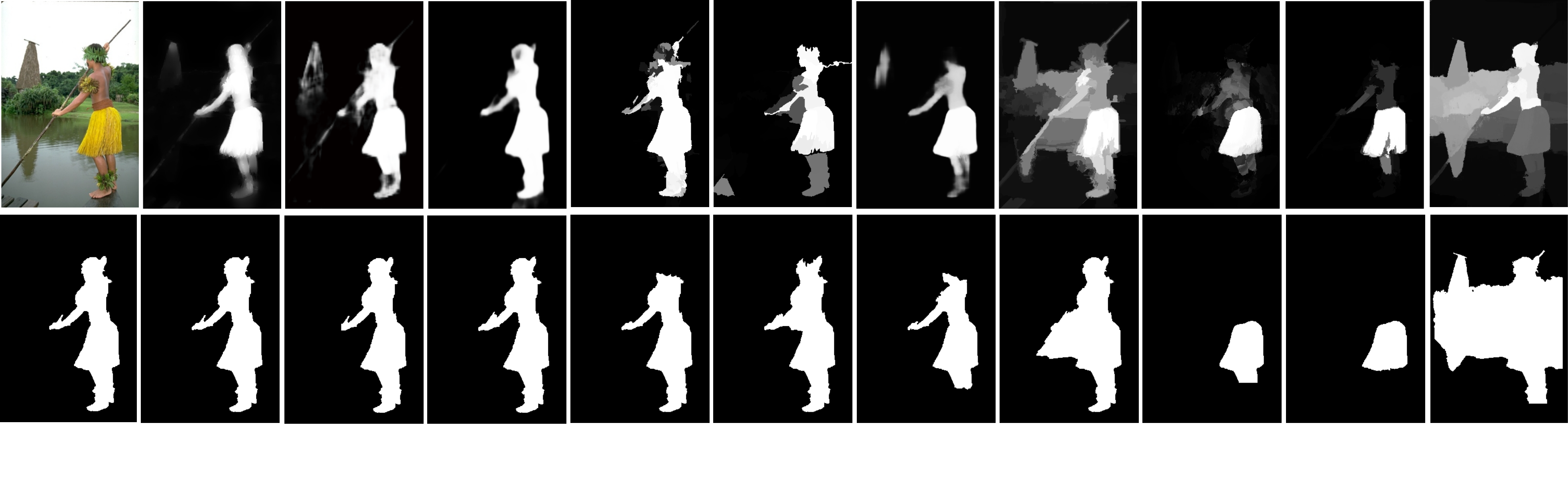} \footnotesize
    \put(3.5,2){GT}
    \put(10.5,2){0.9984}%{0.9994}
    \put(19.5,2){0.9915}%{0.9945}
    \put(28.5,2){0.9867}%{0.9941}
    \put(38.5,2){0.9504}%{0.9567}
    \put(47.5,2){0.9438}%{0.9477}
    \put(56.5,2){0.8951}%{0.8859}
    \put(65.5,2){0.8136}%{0.8119}
    \put(74.5,2){0.6605}%{0.6500}
    \put(83.5,2){0.6358}%{0.6267}
    \put(92.5,2){0.5127}%{0.4525}
    \put(9.5,0){(a)\rfcn}
    \put(18,0){(b)\DISC}
    \put(27.5,0){(c)\DHS}
    \put(37,0){(d)\ELD}
    \put(45.5,0){(e)\MC}
    \put(54.5,0){(f)\DCL}
    \put(63.5,0){(g)\DRFI}
    \put(73.5,0){(h)\DSR}
    \put(82.5,0){(i)\MDF}
    \put(92,0){(j)\ST}
  \end{overpic}
  \caption{Structure-measure ($\lambda=0.25,K=4$) for the outputs of SalCut\cite{cheng2015global}
    algorithm (2nd row) when fed with inputs of 10
    saliency detection algorithms (1st row).
  }\label{fig:damageExample}
   \vspace{-15pt}
\end{figure*}
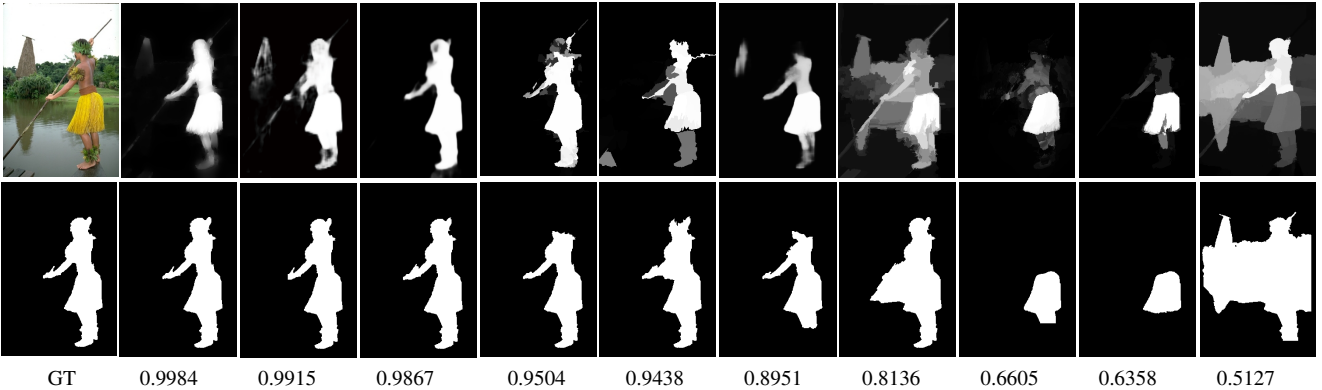

%%%%%%%%% BODY TEXT
\section{Introduction}\label{sec:Introduction}
The evaluation of a predicted foreground map against a ground-truth (GT)
annotation map is crucial in evaluating and comparing various
computer vision algorithm for applications such as
object detection
\cite{borji2015salient,kanan2010robust,bylinskii2015saliency,Qi2015},
saliency prediction
\cite{borji2014salient,DSSalCVPR2017,WangDRFI2017},
image segmentation \cite{qin2014integration},
content-based image retrieval
\cite{ChenPoseShope13,ChengGroupSaliency,JointSalExist17},
semantic segmentation
\cite{hu2013internet,AdversErasingCVPR2017,wei2016stc}
and image collection browsing
\cite{chen2009sketch2photo,li2013partial,ChengSurveyVM2017}.
As a specific example, here we focus on salient object detection models
\cite{ChengSaliency13ICCV,borji2015salient,Borji_PAMI13, borji2015salientDB},
although the proposed measure is
general and can be used for other purposes.
It is necessary to point out that the salient object is not
necessary to be foreground object \cite{feng2016local}.

The GT map is often binary (our assumption here).
The foreground maps are either non-binary or binary.
As a result, evaluation measures can be classified into two types.
The first type is the binary map evaluation
with the common measures being F$_\beta$-measure
\cite{arbelaez2011contour,cheng2015global,liu2011learning}
and PASCAL's VOC segmentation measure \cite{everingham2010pascal}.
The second type is the non-binary map evaluation.
Two traditional measures here include AUC and
AP \cite{everingham2010pascal}.
A newly released measure known as Fbw \cite{2014cvpr/Fbw}
has been proposed to remedy flaws of AP and AUC measures
(see \secref{sec:curntMeasure}).
Almost all salient objection detection methods
output non-binary maps.
Therefore, in this work we focus on non-binary map evaluation.

It is often desired that the foreground map should contain
the entire structure of the object.
Thus, evaluation measures are expected to tell
which model generates a more complete object.
For example, in \figref{fig:FirsCmpDog} (first row) the
blue-border map better captures the dog than the red-border map.
In the latter case, shape of the dog is drastically degraded
to a degree that it is difficult to guess the object category from
its segmentation map.
Surprisingly, all of the current evaluation measures fail
to correctly rank these maps (in terms of preserving the structure).

We employed 10 \sArt saliency detection models to obtain
10 saliency maps (\figref{fig:damageExample}; first row) and
then fed these maps to the SalCut \cite{cheng2015global}
algorithm to generate corresponding binary maps (2th row).
Finally, we used our \textbf{Structure-measure} to rank these maps.
A lower value for our measure corresponds to more
destruction in the global structure of the man (columns e to j).
This experiment clearly shows that our new measure
emphasizes the global structure of the object.
In these 10 binary maps (2rd row),
there are 6 maps with Structure-measure below 0.95,
\ie, with percentage 60\%.
Using the same threshold (0.95),
we found that the proportions of destroyed images
in four popular saliency datasets
(\ie, \ECSSD, \HKU, \PASCAL, and \SOD) are
66.80\%, 67.30\%, 81.82\% and 83.03\%, respectively.
Using the $F_\beta$ measure to evaluate the binary maps,
these proportions are 63.76\%, 65.43\%, 78.32\% and 82.67\%, respectively.
This means that our measure is more restrictive than
the $F_\beta$-measure on object structure.

To remedy the problem of existing measures
(\ie, low sensitivity to global object structure),
we present a structural similarity measure ($\textbf{Structure-measure}$)
\footnote{Source code and results for this measure on the entire
datasets are available at the project page:
\url{http://dpfan.net/smeasure/}.}
based on two observations:
\begin{itemize}
  \item \textbf{Region} perspectives:
    Although it is difficult to describe the object structure of a
    foreground map,
    we notice that the entire structure of an object can be well
    illustrated by combining structures of constituent
    object-parts (regions).
  \item \textbf{Object} perspectives:
  	In the high-quality foreground maps,
  	the foreground region of the maps contrast
    sharply with the background regions and these regions usually have
    approximately uniform distributions.
\end{itemize}

Our proposed similarity measure can be divided into two parts,
including a region-aware structural similarity measure
and an object-aware structural similarity measure.
The region-aware measure tries to capture the global structure information
by combining the structural information of all the object-parts.
The structural similarity of regions has been well explored in the
image quality assessment (IQA) community.
The object-aware similarity measure tries to compare global distributions
of foreground and background regions in SM and GT maps.

We experimentally show that our new measure is more effective than
other measures using 5 meta-measures (a new one introduced by us)
on 5 publicly available benchmark datasets. In the next
section, we review some of the popular evaluation measures.

\section{Current Evaluation Measures}\label{sec:curntMeasure}
Saliency detection models often generate non-binary maps.
Traditional evaluation measures usually convert these non-binary
maps into multiple binary maps.
%In what follows, we describe the process in detail.

%------------------------------------------------------------------
\textbf{Evaluation of binary maps:}
To evaluate a binary map,
four values are computed from the prediction confusion matrix:
True Positives (TP), True Negatives (TN), False Positives (FP)
and False Negatives (FN).
These values are then used to compute three ratios:
True Positive Rate (TPR) or Recall, False Positive Rate (FPR),
and Precision.
The Precision and Recall are combined to compute the
traditional $F_\beta$-measure:
\begin{equation}\label{equ:Fb_measure}
    F_{\beta} =\frac{(1 + \beta^2)Precision \cdot Recall}
      {\beta^2 \cdot Precision + Recall}
\end{equation}

%------------------------------------------------------------------
\textbf{Evaluation of non-binary maps:}
AUC and AP are two universally-agreed evaluation measures.
Algorithms that produce non-binary maps apply three
steps to evaluate the agreement between model predictions
(non-binary maps) and human annotations (GT).
First, multiple thresholds are applied to the non-binary map
to get multiple binary maps.
Second, these binary maps are compared to the binary mask of the GT
to get a set of TPR \& FPR values.
These values are plotted in a 2D plot,
which then the AUC distils the area under the curve.

The AP measure is computed in a similar way.
One can get a Precision \& Recall curve by plotting Precision
$p(r)$ as a function of Recall $r$.
AP measure \cite{everingham2010pascal} is the average value of
$p(r)$ over the evenly spaced x axis points from $r=0$ to $r=1$.

Recently, a measure called Fbw\cite{2014cvpr/Fbw} has offered
an intuitive generalization of the $F_{\beta}$-measure.
It is defined as:
\begin{equation}\label{equ:Fbw_measure}
    F_{\beta}^{\omega} =\frac{(1 + \beta^2)Precision^{\omega}
    \cdot Recall^{\omega}}{\beta^2 \cdot Precision^{\omega} +
    Recall^{\omega}}
\end{equation}
The authors of Fbw identified three causes of inaccurate evaluation of
AP and AUC measures. To alleviate these flaws, they
1) extended the four basic quantities TP, TN, FP, and FN to
non-binary values and,
2) assigned different weights ($w$) to different errors
according to different location and neighborhood information.
While this measure improves upon other measures,
sometimes it fails to correctly rank the foreground maps
(see the 3rd row of the \figref{fig:FirsCmpDog}).
In the next section, we will analyze why the current measures
fail to rank these maps correctly.

%------------------------------------------------------------------
\section{Limitations of Current Measures}

Traditional measures (AP, AUC and Fbw) use four types of basic measures
(FN, TN, FP and TP) to compute Precision, Recall and FPR.
Since all of these measures are calculated in a pixel-wise manner,
the resulting measures (FN, TN, FP and TP) cannot
fully capture the structural information of predicted maps.
Predicted maps with fine structural details are often desired
in several applications.
Therefore, evaluation measures sensitive to foreground structures
are favored.
Unfortunately, the aforementioned measures (AP, AUC and Fbw)
fail to meet this expectation.

\begin{figure}[t!]
  \centering
  \begin{overpic}[width=\columnwidth]{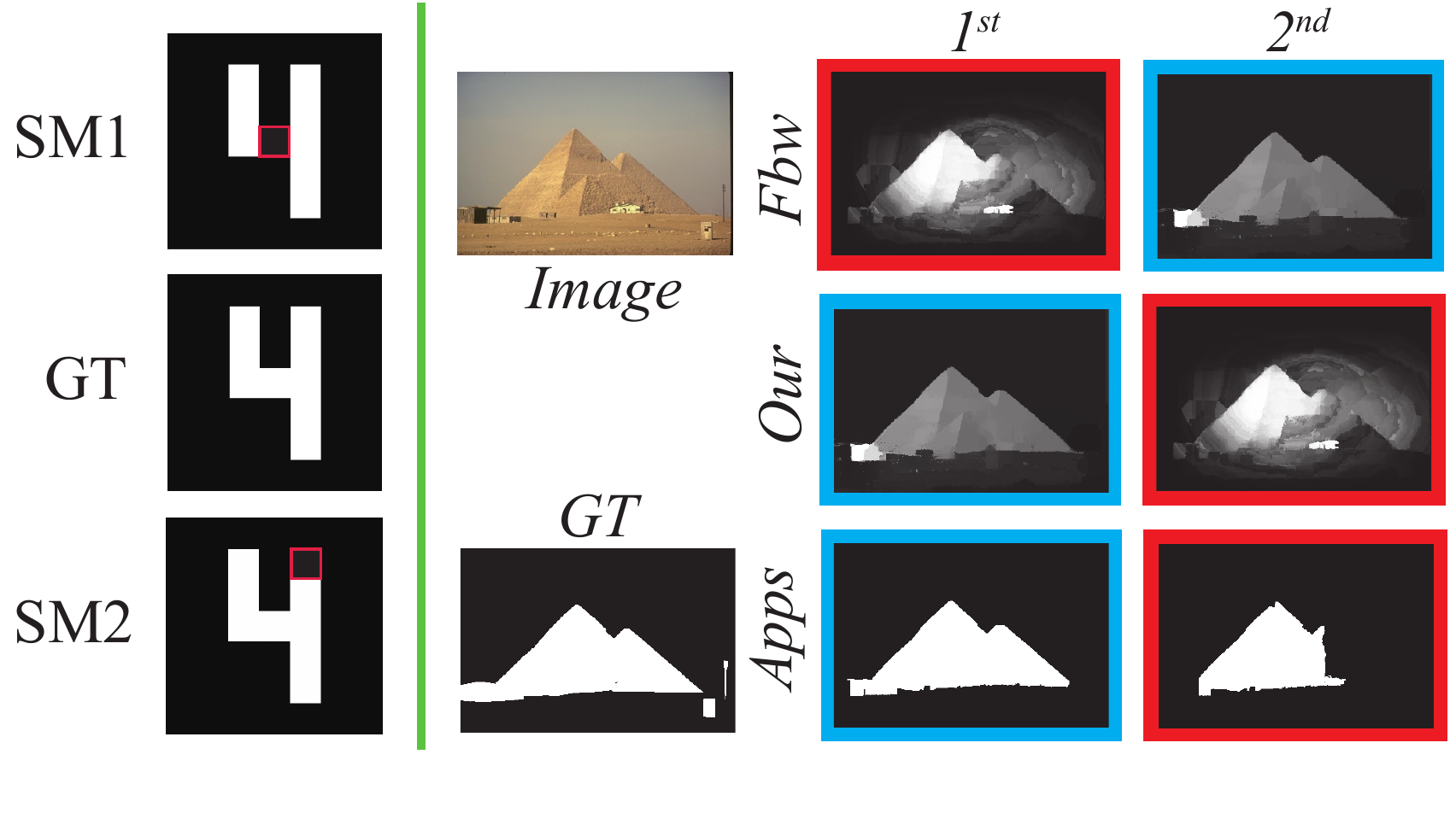}
  \put(10,0){(a)}
  \put(60,0){(b)}
  \end{overpic}
  \caption{\textbf{Structural similarity evaluation}.
    In subfigure (a), two different foreground maps result
    in the same FN, TN, FP, and TP scores.
    In subfigure (b), two maps are produced by two saliency models
    \DSR, and \ST.
    According to the application's ranking and our user-study
    (last row; \secref{sec:ExperimentsApps}),
    the blue-border map is the best, followed by the red-border map.
    Since Fbw measure does not account for the structural similarity,
    it results in a different ranking.
    Our measure (2th row) correctly ranks the blue-border map as higher.
   }\label{fig:cmpAll}
   \vspace{-10pt}
\end{figure}

A typical example is illustrated in \figref{fig:cmpAll} (a)
which contains two different types of foreground maps.
In one, a black square falls inside the digit while
in the other it touches the boundary.
In our opinion, SM2 is favored over SM1 since the latter
destroys the foreground maps more seriously.
However, the current evaluation measures result in the same order.
This is contradictory to our common sense.
%because of the same computed FN, TN, FP and TP.

A more realistic example is shown in \figref{fig:cmpAll} (b).
The blue-border map here better captures the pyramid than
the red-border map,
because the latter offers a fuzzy detection map that mostly highlights
the top part of the pyramid while ignoring the rest.
From an application standpoint
(3th row; the output of the SalCut algorithm fed with saliency maps
and ranked by our measure, \ie, the 2nd row),
the blue-border map offers a complete shape of the pyramid.
Thus, if the evaluation measure cannot capture the object structural
information,
it cannot provide reliable information for model selection
in applications.

%-----------------------------------------------------------------
\section{Our measure}
In this section, we introduce our new measure to
evaluate foreground maps.
In image quality assessment (IQA) field,
a measure known as structural similarity measure ({\sc ssim})
\cite{2004tip/ssim} has been widely used to capture the
structural similarity of the original image and a test image.

Let $x = \{x_{i}|i=1,2,\cdots,N\}$ and $y = \{y_{i}|i=1,2,\cdots,N\}$
be the SM and GT pixel values, respectively.
The $\bar{x}$, $\bar{y}$, $\sigma_{x}$, $\sigma_{y}$ are
the mean and standard deviations of $x$ and $y$.
$\sigma_{xy}$ is the covariance between the two.
Then, SSIM can be formulated as a product of three components:
luminance comparison, contrast comparison and structure comparison.
\begin{equation}\label{equ:ssim}
    ssim =
    \frac{2\bar{x}\bar{y}}{(\bar{x})^{2}+(\bar{y})^{2}}\cdot
    \frac{2\sigma_{x}\sigma_{y}}{\sigma_{x}^{2}+\sigma_{y}^{2}} \cdot
    \frac{\sigma_{xy}}{\sigma_{x}\sigma_{y}}
\end{equation}

In \equref{equ:ssim}, the first two terms denote the luminance comparison
and contrast comparison, respectively.
The closer the two
(\ie, $\bar{x}$ and $\bar{y}$, or $\sigma_{x}$ and $\sigma_{y}$),
the closer the comparison (\ie, luminance or contrast) to 1.
The structures of the objects in an image are independent of the luminance
that is affected by illumination and the reflectance.
So the design of a structure comparison formula should be
independent of luminance and contrast.
{\sc ssim} \cite{2004tip/ssim} associate two unit vectors
$(x-\bar{x})/\sigma_{x}$ and
$(y-\bar{y})/\sigma_{y}$ with the structure of the two images.
Since the correlation between these two vectors is equivalent
to the correlation coefficient between $x$ and $y$,
the formula of structure comparison is denoted by the third
term in \equref{equ:ssim}.

In the field of salient object detection,
researchers are concerned more about the foreground object structures.
%rather than the region similarity.
Thus, our proposed structure measure combines both region-aware
and object-aware structural similarities.
The region-aware structural similarity performs similar to
\cite{2004tip/ssim},
which aims to capture object-part structure information
without any special concern about complete foreground.
The object-aware structural similarity is designed to mainly capture
the structure information of the complete foreground objects.

%------------------------------------------------------------------
\subsection{Region-aware structural similarity measure}
In this section, we investigate how to measure region-aware similarity.
The region-aware similarity is designed to assess the object-part
structure similarity against the GT maps.
We first divide each of the SM and GT maps into four blocks
using a horizontal and a vertical cut-off lines
that intersect at the centroid of the GT foreground.
Then, the subimages are divided recursively like the paper ~\cite{lazebnik2006beyond}.
The total number of blocks is denoted as $K$.
A simple example is shown in \figref{fig:S-measure}.
The region similarity $ssim(k)$ of each block is computed independently
using \equref{equ:ssim}.
We assign a different weight ($w_k$) to each block
proportional to the GT foreground region this block covers.
Thus, the region-aware structural similarity measure can be formulated as
\begin{equation}\label{equ:s_region}
S_r = \sum_{k=1}^{K}{w_k*ssim(k)}
\end{equation}

According to our investigation,
our proposed $S_{r}$ can well describe the object-part similarity
between a SM and a GT map.
We also tried to directly use {\sc ssim} to assess the similarity
between SM and GT at the image level or in the sliding window fashion
as mentioned in \cite{2004tip/ssim}.
These approaches fail to capture region-aware structure similarities.

%------------------------------------------------------------------
\subsection{Object-aware structural similarity measure}
Dividing the saliency map into blocks helps evaluate
the object-part structural similarity.
However, the region-aware measure ($S_{r}$) cannot well account
for the global similarity.
For high-level vision tasks such as salient object detection,
the evaluation of the object-level similarity is crucial.
To achieve this goal, we propose a novel method to assess
the foreground and background separately.
Since, the GT maps usually have important characteristics,
including sharp foreground-background contrast and uniform distribution,
the predicted SM is expected to possess these properties. This helps
easily distinguish foreground from the background.
We design our object-aware structural similarity measure with respect to
these two characteristics.

\textbf{\emph{Sharp foreground-background contrast}}.
The foreground region of the GT map contrasts sharply
with the background region.
We employ a formulation that is similar with the luminance
component of {\sc ssim},
to measure how close the mean probability is between the
foreground region of SM and the foreground region of GT.
Let $x_{FG}$ and $y_{FG}$ represent the probability values
of foreground region of SM and GT, respectively.
$\bar{x}_{FG}$ and $\bar{y}_{FG}$ denote the means of $x_{FG}$
and $y_{FG}$, respectively.
The foreground comparison can be represented as,
\begin{equation}\label{equ:object-fg-sharp}
    O_{FG} = \frac{2\bar{x}_{FG}\bar{y}_{FG}}
    {(\bar{x}_{FG})^{2}+(\bar{y}_{FG})^{2}}.
\end{equation}
\equref{equ:object-fg-sharp} has several satisfactory properties:
\begin{itemize}
\item Swapping the value of $\bar{x}_{FG}$ and $\bar{y}_{FG}$,
$O_{FG}$ will not change the result.
\item The range of $O_{FG}$ is [0,1].
\item If and only if $\bar{x}_{FG}=\bar{y}_{FG}$, we will get $O_{FG}=1$.
\item The most important property, however,
   is that the closer the two maps, the closer the $O_{FG}$ to 1.
\end{itemize}
These properties make \equref{equ:object-fg-sharp}
suitable for our purpose.

\begin{figure}[t!]
\centering
   \begin{overpic}[width=\columnwidth]{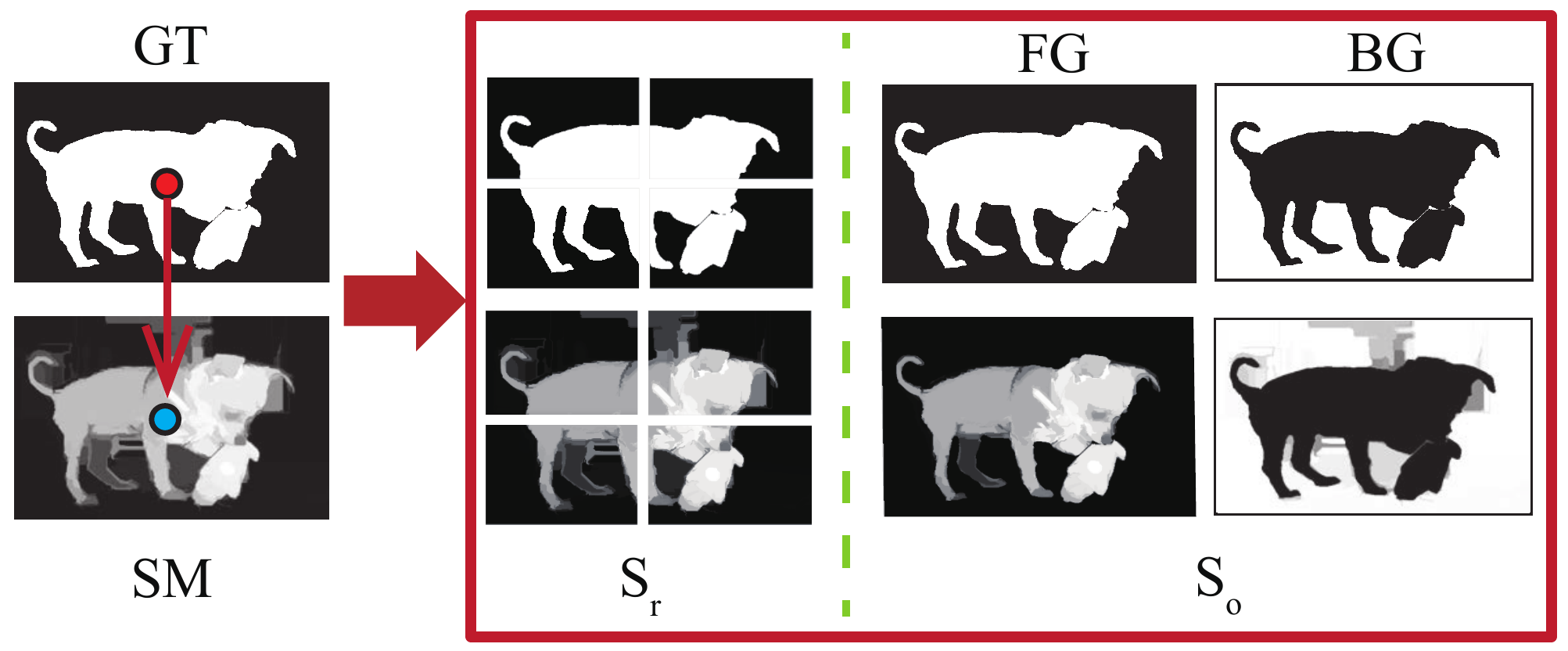}
    \end{overpic}
    \caption{\textbf{Framework of our Structure-measure.}}
    \label{fig:S-measure}
    \vspace{-10pt}
\end{figure}

\textbf{\emph{Uniform saliency distribution}}.
The foreground and background regions of the GT maps usually have
uniform distributions.
So, it is important to assign a higher value to a
SM with salient object being uniformly detected
(\ie, similar saliency values across the entire object).
If the variability of the foreground values in the SM is high,
then the distribution will not be even.

In probability theory and statistics, the coefficient of variation
which is defined as the ratio of the standard deviation to the mean
($\sigma_{x}/\bar{x}$) is a standardized measure of dispersion of a probability distribution.
Here, we use it to represent the dispersion of the SM.
In other words, we can use the coefficient of variation to compute
the distribution of dissimilarity between SM and GT.
According to \equref{equ:object-fg-sharp},
the total dissimilarity between SM and GT
in object level can be written as,
\begin{equation}\label{equ:object-diff}
    D_{FG} = \frac{(\bar{x}_{FG})^{2}+(\bar{y}_{FG})^{2}}
       {2\bar{x}_{FG}\bar{y}_{FG}} +
       \lambda*\frac{\sigma_{x_{FG}}}{\bar{x}_{FG}}
\end{equation}
where $\lambda$ is a constant to balance the two terms.
Since the mean probability of the GT foreground is exactly 1 in practice,
the similarity between SM and GT in object level can be formulated as,
\begin{equation}\label{equ:object-fg}
\begin{aligned}
	O_{FG} = \frac{1}{D_{FG}} = \frac{2\bar{x}_{FG}}{(\bar{x}_{FG})^{2}
    	+1+2\lambda*\sigma_{x_{FG}}}
\end{aligned}
\end{equation}

To compute background comparison $O_{BG}$, we regard the background
as the complementary component of foreground by
subtracting the SM and GT maps
from 1 as shown in \figref{fig:S-measure}.
Then, $O_{BG}$ can be similarly defined as,
\begin{equation}\label{equ:object-fg}
	O_{BG} = \frac{2\bar{x}_{BG}}{(\bar{x}_{BG})^{2}
    	+1+2\lambda*\sigma_{x_{BG}}}
\end{equation}

Let $\mu$ be the ratio of foreground area in GT to image area
($width*height$).
The final object-aware structural similarity measure is defined as,
\begin{equation}\label{equ:S_global}
    S_o = \mu*O_{FG} + (1-\mu)*O_{BG}
\end{equation}

%------------------------------------------------------------------
\subsection{Our new structure-measure}
Having region-aware and object-aware structural
similarity evaluation definitions,
we can formulate the final measure as,
\begin{equation}\label{equ:S-measure}
    S = \alpha*S_o + (1-\alpha)*S_r,
\end{equation}
\noindent where $\alpha\in[0,1]$.
We set $\alpha = 0.5$ in our implementation.
Using this measure to evaluate the three SM maps in
\figref{fig:FirsCmpDog},
we can correctly rank the maps consistent with the application rank.

%------------------------------------------------------------------
\section{Experiments}\label{sec:Experiments}
In order to test the quality of our measure,
we utilized 4 meta-measures proposed by Margolin \etal
\cite{2014cvpr/Fbw}
and 1 meta-measures proposed by us.
These meta-measures are used to evaluate the quality of
evaluation measures \cite{pont2013measures}.
To conduct fair comparisons, all meta-measures are computed on the ASD
(a.k.a ASD1000) dataset \cite{achanta2009frequency}.
The non-binary foreground maps (5000 maps in total) were
generated using five saliency detection
models including \CA, \CB, \RC, \PCA, and \SVO.
We assign $\lambda=0.5$ and $K=4$ in all experiments.
When using a single CPU thread (4 GHz),
our Matlab implementation averagely takes 5.3 ms to
calculate the structure measure of an image.

%------------------------------------------------------------------
\subsection{Meta-Measure 1: Application Ranking}
\label{sec:ExperimentsApps}

An evaluation measure should be consistent with the preferences
of an application that uses the SM as input.
We assume that the GT map is the best for the applications.
Given a SM, we compare the application's output
to that of the GT output.
The more similar a SM is to the GT map,
the closer its application's output should be to the GT output.

To quantify the accuracy in ranking, we use the SalCut
\cite{cheng2015global} as the application to perform
this meta-measure.

\begin{figure}[t!]
    \begin{overpic}[width=\columnwidth]{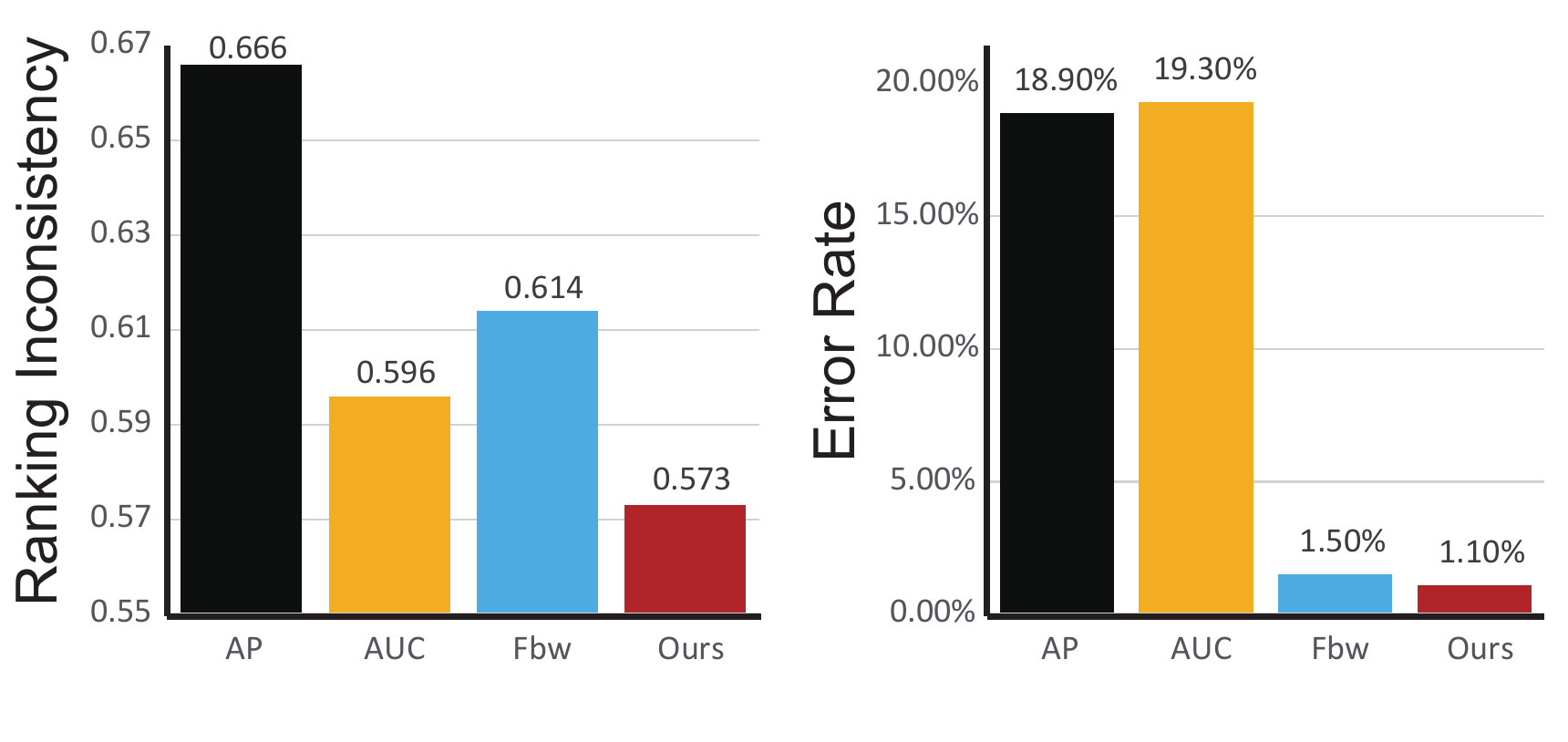}
     \put(10,0){(a) Meta-Measure 1}
     \put(60,0){(b) Meta-Measure 2}
    \end{overpic}
    \caption{\textbf{Meta-measure 1\&2-results.}}
    \label{fig:metaMeasure1_2}
    \vspace{-10pt}
\end{figure}

Here, we utilize 1-Spearman's $\rho$ measure \cite{best1975algorithm}
to evaluate the ranking accuracy of the measures,
where lower values indicates better ranking consistency.
Comparison between different measurements are shown in
\figref{fig:metaMeasure1_2} (a),
which indicates that our structure measure produces best ranking consistency
among other alternative methods.

%------------------------------------------------------------------
\subsection{Meta-Measure 2: State-of-the-art vs. Generic}
The second meta-measure is that a measure should prefer
the output achieved by a \sArt method over generic baseline maps
(\eg, centered Gaussian map) that discard the image content.
A good evaluation measure should rank the SM generated by a
\sArt model higher than a generic map.

We count the number of times a generic map scored higher than
the mean score generated by the five \sArt models
(\CA, \CB, \RC, \PCA, \SVO).
The mean score provides an indication of model robustness.
%to cases in which a specific models.
The results are shown in \figref{fig:metaMeasure1_2} (b).
The lower the value here, the better.
Over  1000 images, our measure has only 11 errors
(\ie, generic winning over the s.t.a).
Meanwhile, the AP and AUC measures are very poor and make
significantly more mistakes.

%------------------------------------------------------------------
\subsection{Meta-Measure 3: Ground-truth Switch}
The third meta-measure specifies that a good SM should not
obtain a higher score when switching to a wrong GT map.
In Margolin \etal \cite{2014cvpr/Fbw}, a SM is considered as ``good''
when it scores at least 0.5 out of 1
(when compared to the original GT map).
Using this threshold (0.5), top 41.8\% of the total
5000 maps were selected as ``good'' ones.
For a fair comparison, we
follow Margolin \etal to select the same percentage of ``good'' maps.
For each of the 1000 images, 100 random GT switches were tested.
We then counted the percentage of times that a measure increases a
saliency map's score when an incorrect GT map was used.

\begin{figure}[t!]
    \begin{overpic}[width=\columnwidth]{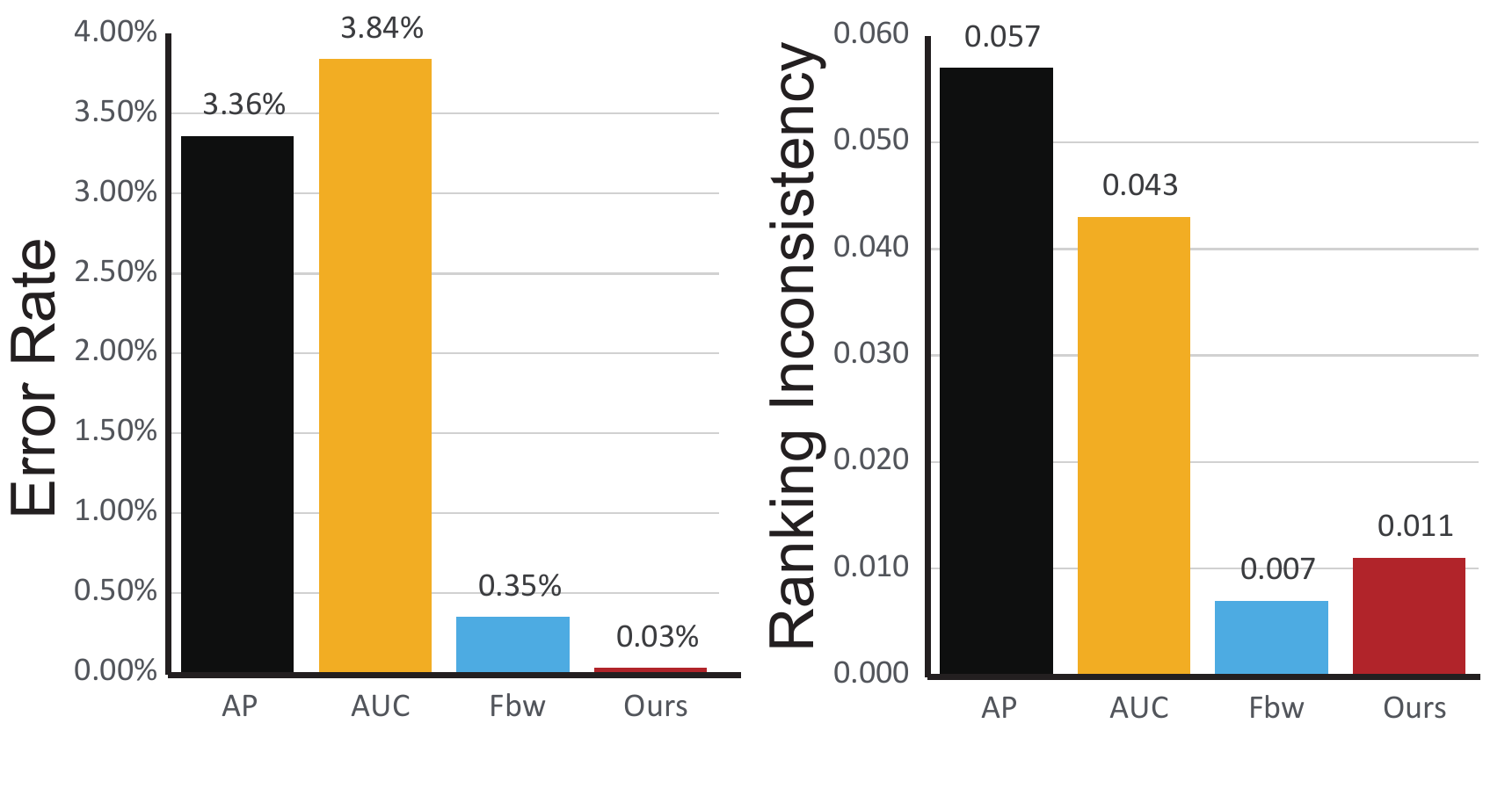}
     \put(10,0){(a) Meta-Measure 3}
     \put(60,0){(b) Meta-Measure 4}
    \end{overpic}
    \caption{\textbf{Meta-measure 3\&4-results.}}
    \label{fig:metaMeasure3_4}
 % \vspace{-5pt}
\end{figure}

The \figref{fig:metaMeasure3_4} (a) shows the results.
The lower the score, the higher capability to match to the correct GT.
Our measure performs the best about 10 times better
compared to the second best measure.
This is due to the fact that our measure captures the
object structural similarity between a SM and a GT map.
Our measure will assign a lower value to the ``good'' SM
when using a random selected GT since the object structure
has changed in the random GT.

\begin{figure}[t!]
\centering
    \begin{overpic}[width=\columnwidth]{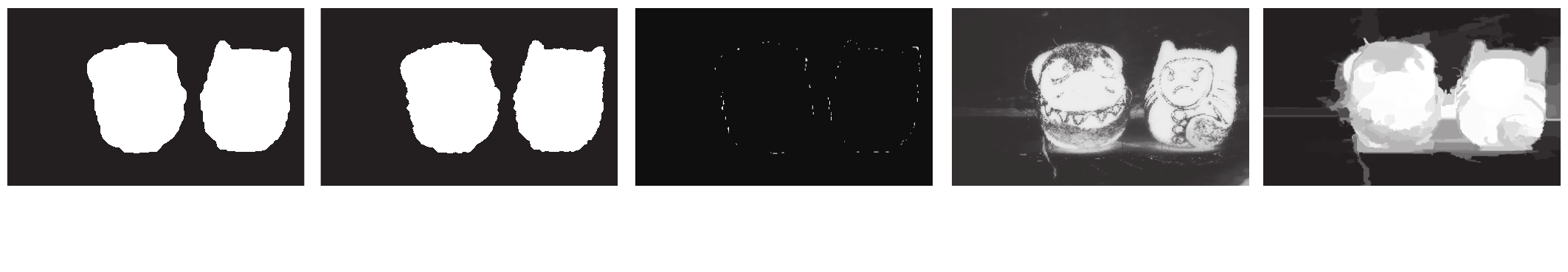}
       \put(1,1){(a) GT}
       \put(23,1){(b) GT}
       \put(43,1){(c) Dif}
       \put(61,1){(d) SM1}
       \put(81,1){(e) SM2}
    \end{overpic}
    \caption{\textbf{Meta-measure 4: Annotation errors.}
    (a) ground-truth map,
    (b) morphologically changed version of a,
    (c) difference map between a and b,
    (d) saliency map1,
    (e) saliency map2.
    }\label{fig:AnnotationError}
    \vspace{-5pt}
\end{figure}
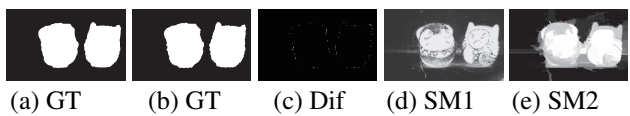

%-----------------------------------------------------------------
\subsection{Meta-Measure 4: Annotation errors}
The fourth meta-measure specifies that an evaluation measure
should not be sensitive to slight errors/inaccuracies in the manual
annotation of the GT boundaries.
To perform this meta-measure, we make a slightly modified
GT map by using morphological operations.
An example is shown in \figref{fig:AnnotationError}.
While the two GT maps in (a) \& (b) are almost identical,
measures should not switch the ranking between the two saliency
maps when using (a) or (b).

We use 1-Spearman's Rho measure to examine
the ranking correlation before and after the annotation errors were
introduced. The lower the score, the more robust an evaluation measure
is to annotation errors~\cite{2014cvpr/Fbw}.
The results are shown in \figref{fig:metaMeasure3_4} (b). Our measure
outperforms both the AP and the AUC but not the best.
Inspecting this finding, we realized that it is not always the case
that the lower the score, the better an evaluation measure. The reason
is that sometimes ``slight" inaccurate manual annotations can change the
structure of the GT map, which in turn can change the rank.
We examined the effect of structure change carefully. Major structure
change often corresponds to continuous large regions in the difference
map between GT and its morphologically changed version.
We try to use the sum of corroded version of the difference map as
measure of major structure change and sort all GT maps.

\begin{figure}[t!]
    \centering
    \begin{overpic}[width=\columnwidth]{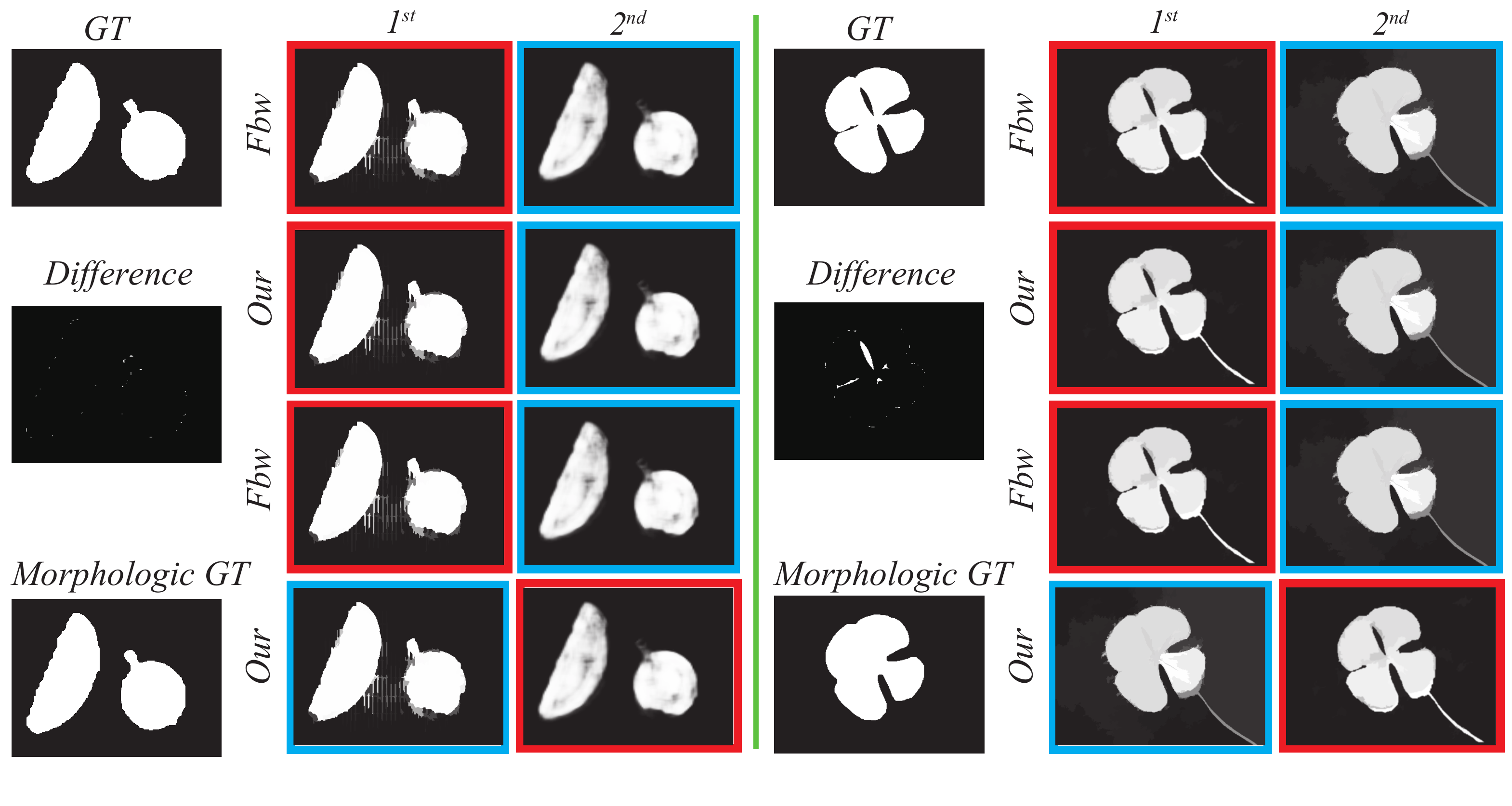}
    \put(5,0){(a) Structural unchanged}
    \put(55,0){(b) Structural changed}
    \end{overpic}
    \caption{\textbf{Structural unchanged/changed.}
      (a) Both of our and Fbw measures are not sensitive to a
      inaccuracies (structural unchanged) in the manual annotation of
      the GT boundaries.
      (b) The ranking of an evaluation measure should be sensitive to
      the structural changes. Surprisingly, the current best measure
      (Fbw) cannot adaptive to the structural changes. Using our
      measure, we can change the rank correctly.
      Best viewed on screen.
    }\label{fig:meta-measure4example}
   \vspace{-10pt}
\end{figure}

Among top 10\% least change GT maps, our measure and Fbw have the
same MM4 scores (same rank).
When the topology of GT map does not change,
our measure and Fbw measure keep the original ranking.
We can see from the example \figref{fig:meta-measure4example} (a).
While ground truth maps (GT and Morphologic GT) differ slightly,
both Fbw and our measure keep the ranking order of the two saliency maps,
depending on the GT used.

\begin{figure}
  \centering
  \includegraphics[width=\linewidth]{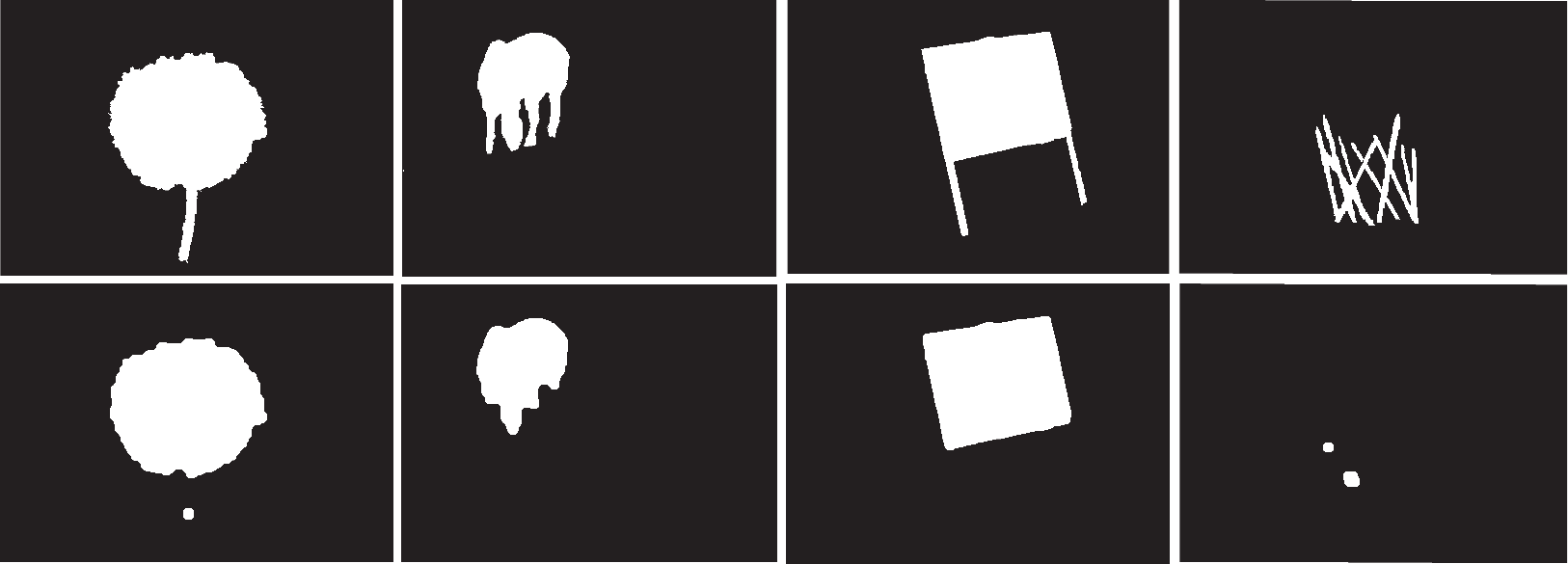}
  \caption{\textbf{Structural changes examples.}
     The first row are GT maps.
     The second row are its morphologically changed version.
  }\label{fig:Morphological-GT}
\end{figure}

\begin{table*}[t]
	\centering
	\footnotesize
	\renewcommand{\tabcolsep}{1.5mm}
    \caption{Quantitative comparison with current measures on 3 meta-Measures.
      The best result is highlighted in {\textbf{\color{blue}{blue}.}
      MM:meta-Measure}.
    }\label{table:3_Dataset}
    \vspace{5pt}
    \begin{tabular*}{\textwidth}{l|ccc|ccc|ccc|ccc} \hline
        \multirow{2}*{} &  & \PASCAL & &  & \ECSSD & &  & \SOD &  &  & \HKU & \\ \cline{2-13}
        ~~~~~~~~~~~~~& MM1 &  MM2(\%) &  MM3(\%) & MM1 & MM2(\%) & MM3(\%) & MM1 & MM2(\%) & MM3(\%) & MM1 & MM2(\%) & MM3(\%) \\ \hline
        AP  & 0.452 & 12.1 & 5.50 & 0.449 & 9.70 & 3.32 & 0.504 & 9.67 & 7.69 & 0.518 & 3.76 & 1.25 \\
        AUC &
              0.449 & 15.8 & 8.21 & 0.436 & 12.1 & 4.18 & 0.547 & 14.0 & 8.27 & 0.519 & 7.02 & 2.12 \\
        Fbw &
              0.365 & 7.06 & 1.05 & 0.401 & \textbf{\color{blue}{3.00}} & 0.84 & 0.384 & 16.3 & 0.73 & 0.498 & 0.36 & 0.26 \\
        $\text{Ours}$ &
        \textbf{\color{blue}{0.320}}  & \textbf{\color{blue}{4.59}} & \textbf{\color{blue}{0.34}} & \textbf{\color{blue}{0.312}} & 3.30 & \textbf{\color{blue}{0.47}} & \textbf{\color{blue}{0.349}} & \textbf{\color{blue}{9.67}} & \textbf{\color{blue}{0.60}} & \textbf{\color{blue}{0.424}} & \textbf{\color{blue}{0.34}} & \textbf{\color{blue}{0.08}} \\ \hline
    \end{tabular*}
\end{table*}

\begin{figure*}[t!]
\vspace{-10pt}
  \begin{overpic}[width=\textwidth]{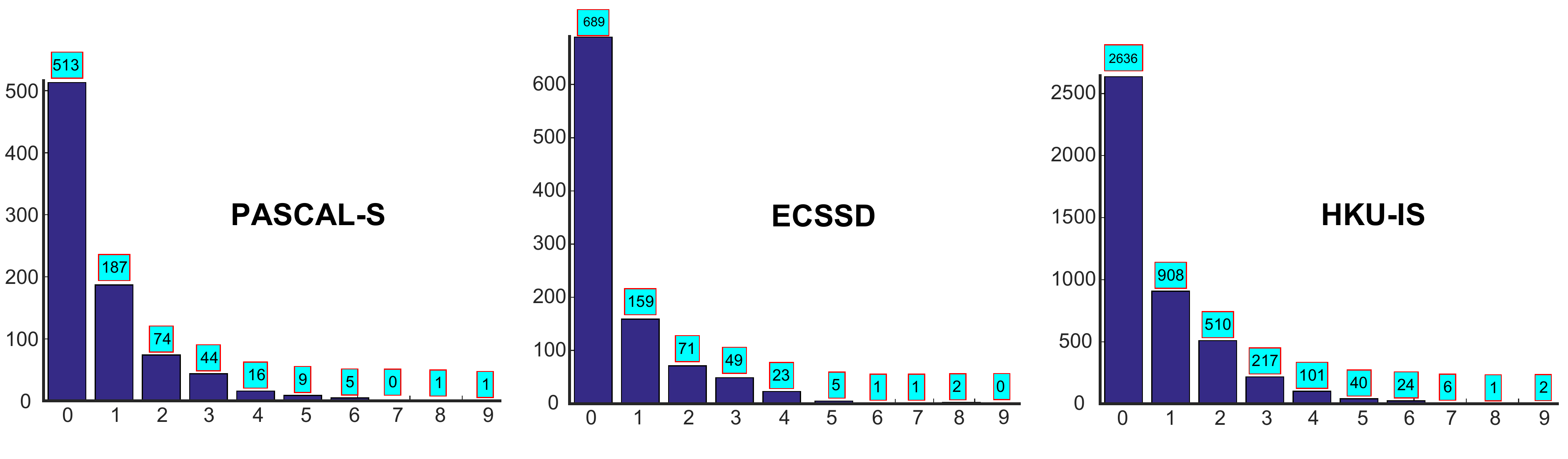} \small
    \put(15,0){(a)}
    \put(50,0){(b)}
    \put(85,0){(c)}
  \end{overpic}
  \caption{\textbf{The rank distance between Fbw and our measure.}
    The (a)-(c) is the three datasets that present
    the rank distance between Fbw and our \textbf{Structure-measure}.
    The y axis of the plot is the number of the images.
    The x axis is the rank distance.
  }\label{fig:User-StudyDataset}
  \vspace{-10pt}
\end{figure*}

Among top 10\% most changes GT maps, we asked 3 users to judge whether
the GT maps have major structure change. 95 out of 100
GT maps were considered to have major structure change,
(similar to \figref{fig:Morphological-GT}, such as small bar, thins legs,
slender foot and minute lines in each group),
for which we believe that keeping rank stability is not good.
\figref{fig:meta-measure4example} (b) demonstrates this argument.
When we use the GT map as the reference, Fbw and our measure rank
the two maps properly. However, when using Morphologic GT as the reference,
ranking results are different. Clearly, the blue-border SM is visually and
structurally more similar to the Morphologic GT map than the red-border SM.
The measure should rank the blue-border SM higher than red-border SM. So the
ranking of these two maps should be changed. While the Fbw measure fails
to meet this end, our measure gives the correct order.

Above-mentioned analysis suggests that this meta-measure is not very reliable.
Therefore, we do not include it in our further comparison on other datasets.

%------------------------------------------------------------------
\subsection{Further comparison}

The results in \figref{fig:metaMeasure1_2} \& \figref{fig:metaMeasure3_4} (a)
show that our measure achieves the best performance using 3 meta-measures
over the ASD1000 dataset. However, a good evaluation measure should perform
well over almost all datasets. To demonstrate the robustness of our measure,
we further performed experiments on four widely-used benchmark datasets.

\textbf{Datasets.}
The used datasets include \PASCAL, \ECSSD, \HKU, and \SOD. PASCAL-S
contains $850$ challenging images,
which have multiple objects with high background clutter.
ECSSD contains $1000$ semantically meaningful
but structurally complex images.
HKU-IS is another large dataset that contains $4445$ large-scales images.
Most of the images in this dataset contain more than one salient
object with low contrast.
Finally, we also evaluate our measure over SOD dataset,
which is a subset of the BSDS dataset.
It contains a relatively small number  of images (300),
but with multiple complex objects.

\textbf{Saliency Models.}
We use 10 \sArt models including 3 traditional
models (\ST, \DRFI, and \DSR ) and  7 deep learning based models
(\DCL, \rfcn, \MC, \MDF, \DISC, \DHS, and \ELD) to test the measures.

\textbf{Results.}
Results are shown in \tabref{table:3_Dataset}.
Our measure performs the best according to the first meta-measure.
This indicates that our measure is more useful for applications than others.
According to meta-measure 2, our measure performs better
than the existing measures, except that ECSSD where it
is ranked second.
For meta-measure 3, our measure reduces the error rate by
$67.62\%$, $44.05\%$, $17.81\%$, $69.23\%$ in PASCAL, ECSSD,
SOD and HKU-IS, respectively compared to the second ranked measure.
This indicates that our measure has higher capacity
to measure the structural similarity between a SM and a GT map.
All in all, our measure wins in the majority of cases
which clearly demonstrates that our new measure is
more robust than other measures.

%------------------------------------------------------------------
\subsection{Meta-Measure 5: Human judgments}
Here, we propose a new meta-measure to evaluate
foreground evaluation measures.
This meta-measure specifies that the map ranking according
to an evaluation measure should agree with the human ranking.
It is argued that ``a human being is the best judge to evaluate
the output of any segmentation algorithm'' \cite{pal1993review}.
However, subjective evaluation over all images of a dataset is
impractical due to time and monetary costs.
To the best of our knowledge, there is no such visual similarity
evaluation database available that meets these requirements.

\textbf{Source saliency maps collection.}
The source saliency maps are sampled from the three large scale datasets:
PASCAL-S, ECSSD, and HKU-IS.
As mentioned above, we use 10 \sArt saliency models to
generate the saliency maps in each dataset.
Therefore, we have 10 saliency maps for each image.
We use Fbw and our measure to evaluate the 10 maps and
then pick the first ranked map according to each measure.
If the two measures choose the same map,
their rank distance is 0.
If one measure ranks a map first,
but the other ranks the same map in the $n$-th place,
then their rank distance is $|n-1|$.
\figref{fig:User-StudyDataset} (a), (b) and (c)
show the rank distance between the two measures (\ie, histogram).
The blue-box is the number of images for each rank distance.
Some maps with rank distance greater than 0 are chosen as
candidates for our user study.

\begin{figure}[t!]
\vspace{-10pt}
\centering
    \begin{overpic}[width=\columnwidth]{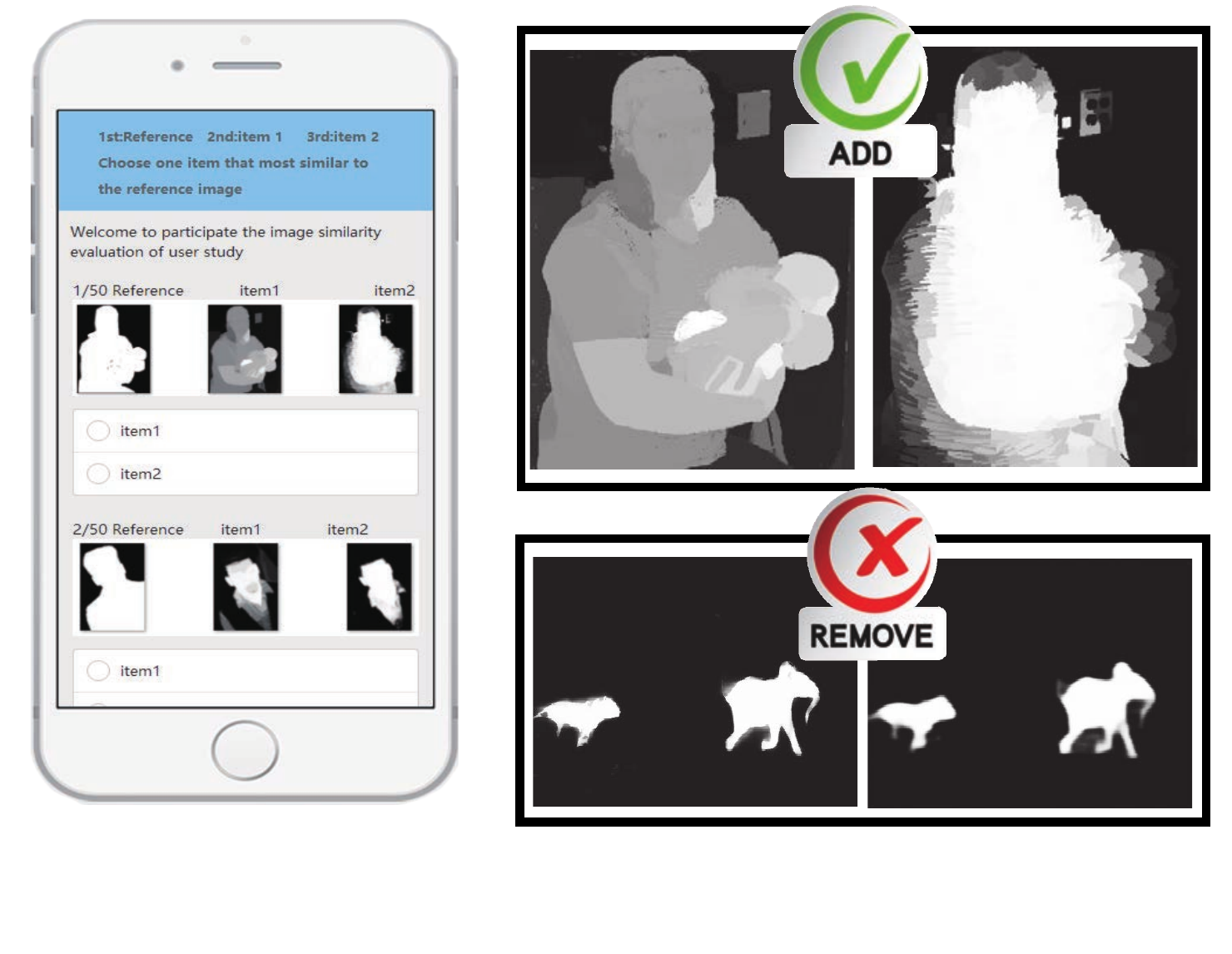}
    \put(20,3){(a)}
    \put(65,3){(b)}
    \end{overpic}
    \caption{\textbf{Our user study platform}.}
    \label{fig:UserApps}
    %\vspace{-5pt}
\end{figure}

\begin{figure}[t!]
%\vspace{-10pt}
\centering
    \begin{overpic}[width=\columnwidth]{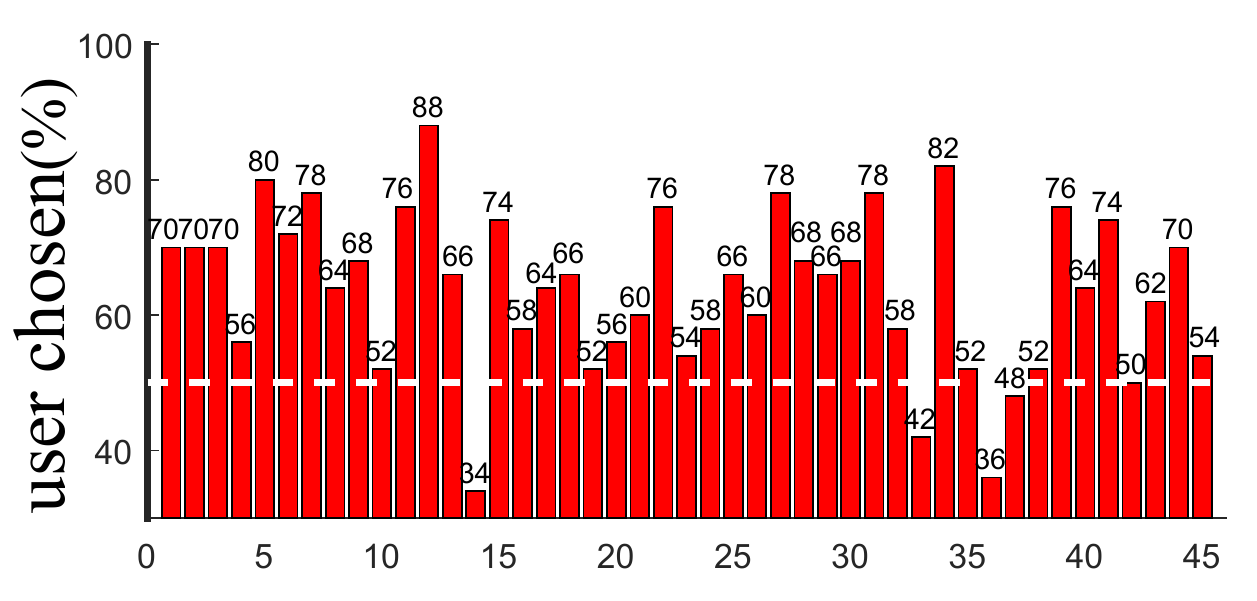}
    \end{overpic}
    \caption{\textbf{Results of our user study}.
        The x axis is the viewer id.
        The y axis shows the percentage of the trials in
        which a viewer preferred the map chosen by our measure.}
    \label{fig:User-Study-Result}
    \vspace{-5pt}
\end{figure}

\textbf{User study.}
We randomly selected 100 pairs of maps from the three datasets.
The top panel in \figref{fig:UserApps} (b) shows
one example trial where
the best map according to our measure in the left,
and the best map according to the Fbw on the far right.
The user is asked to choose the map he/she
thinks resembles the most with the GT map.
In this example, these two maps are obviously different
making the user decide easily.
In another example (bottom panel in \figref{fig:UserApps} (b)),
the two maps are too similar making it difficult to
choose the one closet to the GT.
Therefore, we avoid showing such cases to the subjects.
Finally, we are left with a stimulus set of size 50 pairs.
We developed a mobile phone app to conduct the user study.
We collected data from 45 viewers who were naive to
the purpose of the experiment.
Viewers had normal or corrected vision.
(Age distribution is 19-29 years old; Eduction from undergraduate to Ph.D;
10 different major such as history, medicine and finance;
25 males and 20 females)

\textbf{Results.}
Results are shown in \figref{fig:User-Study-Result}.
The percentage of trials (averaged over subjects) in which
a viewer preferred the map chosen by our measure is $63.69\%$.
We used the same way to do another 2 user study experiments
( AP compare to our measure, AUC compare to our measure).
The results are $72.11\%$ and $73.56\%$ respectively,
which means that our measure correlates better with human judgments.

\subsection{Saliency model comparison}
\begin{figure}[t!]
\vspace{-10pt}
\centering
    \begin{overpic}[width=.95\columnwidth]{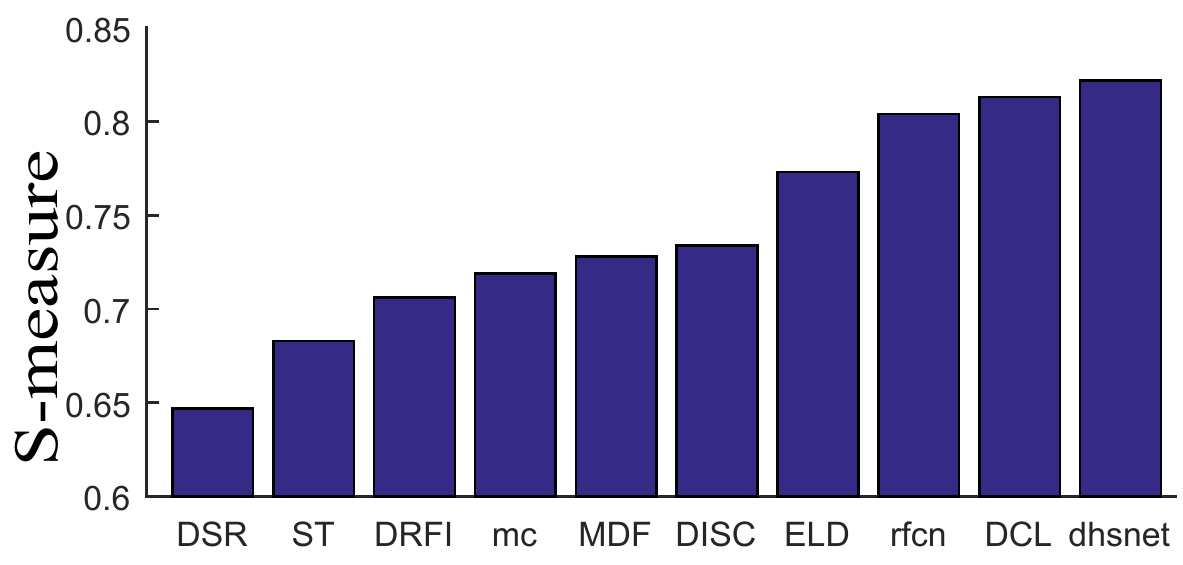}
    \end{overpic}
    \caption{\textbf{Ranking of 10 saliency models using our new
    measure.} The y axis shows the average score on each dataset
    (\PASCAL, \ECSSD, \HKU, \SOD).
    }\label{fig:SmeasureRankBar}
    \vspace{-5pt}
\end{figure}

Establishing that our Structure-measure offers a better way to
evaluate salient object detection models,
here we compare 10 \sArt saliency models on
4 datasets (PASCAL-S, ECSSD, HKU-IS, and SOD).
\figref{fig:SmeasureRankBar} shows the rank of 10 models.
According to our measure, the best models in order are dhsnet,
DCL and rfcn.
Please see the supplementary material for sample maps of these models.

%-------------------------------------------------------------------
\section{Discussion and Conclusion}
In this paper, we analyzed the current saliency evaluation measures
based on pixel-wise errors and showed that
they ignore the structural similarities.
We then presented a new structural similarity measure
known as $\textbf{Structure-measure}$
which simultaneously evaluates region-aware and object-aware structural similarities between a saliency map and a ground-truth map.
Our measure is based on two important characteristics:
1) sharp foreground-background contrast, and
2) uniform saliency distribution.
Further, the proposed measure is efficient and easy to calculate.

Experimental results on 5 datasets
demonstrate that our measure performs better than the current measures
including AP, AUC, and Fbw.
Finally, we conducted a behavioral judgment study over a
database of 100 saliency maps and 50 GT maps.
Data from 45 subjects shows that on average they
preferred the saliency maps chosen by our measure
over the saliency maps chosen by the AP, AUC and Fbw.

In summary, our measure offers new insights into
salient object detection evaluation where current measures
fail to truly examine the strengths and weaknesses of saliency models.
We encourage the saliency community to consider this measure
in future model evaluations and comparisons.

\vspace{-5pt}
\paragraph{Acknowledgement}
We would like to thank anonymous reviewers for
their helpful comments on the paper.
This research was supported by NSFC (NO. 61572264, 61620106008),
Huawei Innovation Research Program, CAST YESS Program,
and IBM Global SUR award.

{\small
\bibliographystyle{ieee}
\bibliography{egbib}
}

\end{document}